\documentclass[10pt,twocolumn,letterpaper]{article}

\usepackage[pagenumbers]{cvpr} 
\usepackage{bm}
\usepackage[dvipsnames]{xcolor}

\definecolor{cvprblue}{rgb}{0.21,0.49,0.74}
\usepackage[pagebackref,breaklinks,colorlinks,citecolor=cvprblue]{hyperref}
\usepackage{booktabs}
\usepackage{multirow}
\usepackage{float}
\usepackage{setspace}
\def\paperID{4198} 
\def\confName{CVPR}
\def\confYear{2024}

\title{ADFactory: An Effective Framework for Generalizing Optical Flow with Nerf }
\author{Han Ling, Quansen sun, Xian Xu\\
NJUST\\
{\tt\small 320106010190@njust.edu.cn}
}

\begin{document}

\twocolumn[{%
	\maketitle
	\vspace{-3.0em}
	\begin{figure}[H]
		\hsize=\textwidth 
		\centering
		\includegraphics[width=6.8in]{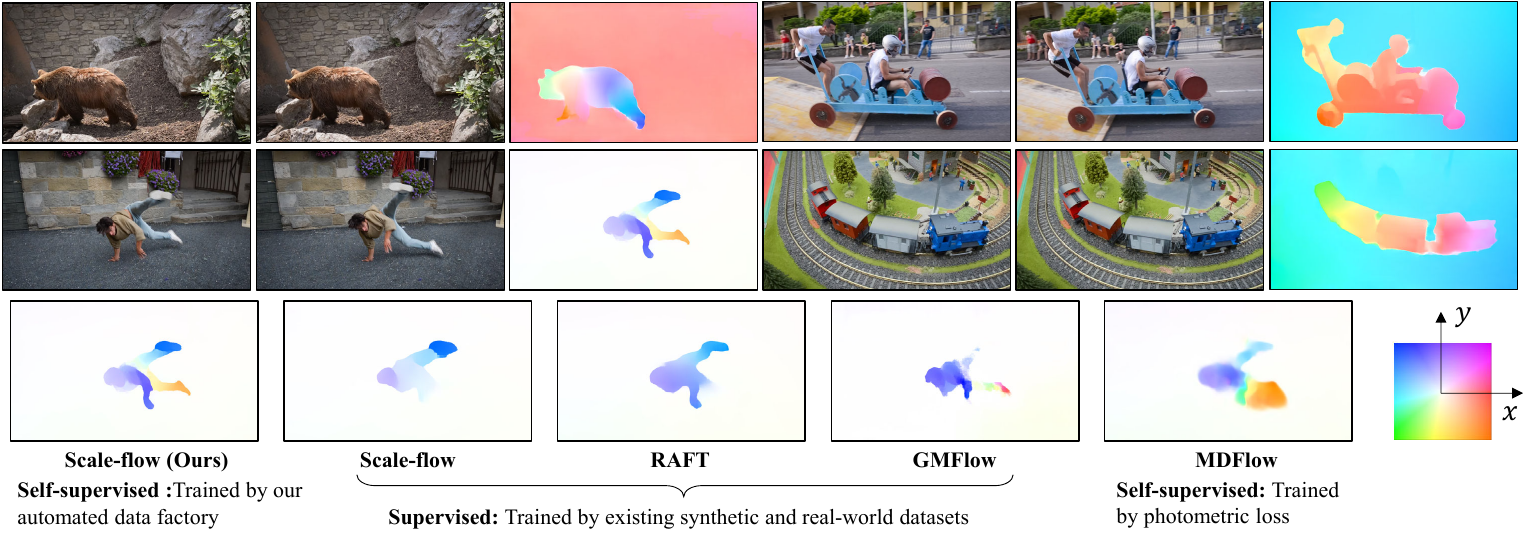}
		\caption{\textbf{Zero-Shot Generalization Results in Real World.} On top is Scale-flow using our data factory scheme to estimate optical flow results in real-world scenarios. Below is a comparison with existing advanced supervised and self-supervised methods. Our method shows unprecedented accuracy and clarity. Moreover, our fully automated data factory requires no manual intervention and only utilizes photos captured by a monocular camera to train optical flow tasks.}
		\label{fig_top}
	\end{figure}}]

\begin{abstract}

A significant challenge facing current optical flow methods is the difficulty in generalizing them well to the real world.
This is mainly due to the high cost of hand-crafted datasets, and existing self-supervised methods are limited by indirect loss and occlusions, resulting in fuzzy outcomes.
To address this challenge, we introduce a novel optical flow training framework: automatic data factory (ADF). ADF only requires RGB images as input to effectively train the optical flow network on the target data domain.
Specifically, we use advanced Nerf technology to reconstruct scenes from photo groups collected by a monocular camera, and then calculate optical flow labels between camera pose pairs based on the rendering results.
To eliminate erroneous labels caused by defects in the scene reconstructed by Nerf, we screened the generated labels from multiple aspects, such as optical flow matching accuracy, radiation field confidence, and depth consistency. The filtered labels can be directly used for network supervision.
Experimentally, the generalization ability of ADF on KITTI surpasses existing self-supervised optical flow and monocular scene flow algorithms. In addition, ADF achieves impressive results in real-world zero-point generalization evaluations and surpasses most supervised methods. Code will be available.

\end{abstract}    
\section{Introduction}

Optical flow aims to estimate the motion of each pixel on the image plane between two consecutive frames. It is one of the oldest problems in computer vision and has essential applications in autonomous driving\cite{capito2020optical,wang2021end,wang2021learning}, video understanding\cite{fan2018end}, and human action recognition\cite{lertniphonphan2011human,arshad2022human}. In recent years, deep learning methods\cite{sun2018pwc,teed2020raft} have become the mainstream solution for optical flow problem and have achieved performance far superior to traditional handcrafted methods \cite{beauchemin1995computation} in benchmark tests\cite{Menze2015ISA,butler2012naturalistic}.

Compared to traditional methods, deep optical flow methods can learn robust matching features and contextual features from large-scale data, as well as infer the optical flow of occluded parts\cite{jiang2021learning}. However, acquiring real-world optical flow datasets is challenging. It requires additional sensors such as LiDAR, GPS, IMU, and much manual annotations\cite{Menze2015ISA}, which has been one of the significant limitations of applying optical flow methods\cite{xu2022gmflow}. Although recent works in synthetic datasets\cite{mayer2016large,butler2012naturalistic,mehl2023spring} and self-supervised learning\cite{jonschkowski2020matters,kong2022mdflow} have alleviated this limitation to some extent, they are still subject to constraints in domain transferability, lighting variations, and occlusions.

In this era where data is the gold mine, massive training datasets are a necessary factor for the success of powerful models such as ChatGPT\cite{brown2020language} and SAM\cite{kirillov2023segment}. 
However, how to quickly and cost-effectively mine high-quality training data from the `gold mines' and fully tap into the enormous potential of deep networks remains an unexplored challenge.
In this paper, we propose a novel automated data factory (ADF) to address this issue.  ADF utilizes only photos taken by a single camera as input, and does not require any manual intervention or ground-truth labels. It can quickly generate almost infinite datasets at a meagre cost, making it a valuable asset for various geometric matching tasks. Specifically, ADF mainly consists of two parts: data generation and data filtering.

\textbf{Data generation:} We build a data generator based on the latest radiation rendering technology (Nerf)\cite{barron2023zip}. Firstly, based on Nerf, a high-resolution scene is reconstructed from photos taken by a monocular camera. Then, by rendering a depth map and randomly generating new camera pose pairs, we can calculate the corresponding optical flow results between poses. In addition, to alleviate the problem that Nerf can only reconstruct static scenes with high quality, we also added randomly shaped foreground floaters based on the Bezier curve.

\textbf{Data filtering:}
In Nerf reconstruction, it is inevitable to encounter areas of reconstruction failure, resulting in further generation of erroneous training data. Therefore, constructing a comprehensive data filtering mechanism is essential.
Before us, the only evaluation metric used was ambient occlusion (AO) \cite{truong2023sparf,tosi2023nerf}. However, AO can only be used to check whether floating objects are between the viewpoint and the target depth, which cannot be widely used for Nerf methods and downstream tasks.
In this paper, we have designed practical evaluation metrics such as reconstruction stability, depth consistency, and optical flow visual consistency based on the characteristics of the Nerf method. Experiments have shown that the combination of these metrics improves the overall performance of the optical flow method by at least 20\%.


The filtered optical flow results can be directly used to train complex geometric matching tasks\cite{sun2021loftr,xu2022gmflow,liu2022camliflow,ling2023learning}. As shown in Fig.\ref{fig_top}, the optical flow model trained on ADF-generated data achieves stunning real-world generalization effects.
We believe that ADF is an important step in making optical flow methods widely applicable in real-world scenarios and a stepping stone to building matching large models. In this paper, we only used about 300 collected scenarios, which has made the ADF-supervised optical flow methods far superior in zero-shot generalization to self-supervised and supervised models trained on existing datasets \cite{mayer2016large,butler2012naturalistic,Menze2015ISA}.

Our contributions can be summarized as follows:

\begin{itemize}
	\item{A new mode for self-supervised learning of optical flow algorithms based on Nerf.}
	\item{A fully automatic optical flow label generation pipeline ADF, which only requires images taken by a monocular camera as input to generate almost infinite datasets.}
	\item{We collected and created an optical flow dataset ADF58 containing over 300 scenes and 58800 images. Experimental results demonstrate that ADF58 achieves superior zero-shot generalization capability in optical flow tasks compared to existing datasets.}

\end{itemize}

\section{Preliminaries}
\label{sec:pre}
In this section, we mainly introduce the components of the Nerf method used in this paper and analyze how to render the most critical depth values.

We use the Zip-Nerf\cite{barron2023zip} to represent 3D scenes and image generation. This is currently one of the most advanced Nerf methods in terms of performance, which can achieve anti-aliasing and robust reconstruction results. Secondly, it uses hash voxel networks to store color and density features. This explicit expression not only accelerates convergence speed, but also naturally conforms to multi-view geometric structure consistency better than implicit expression.

\textbf{Scene representation:} Zip-Nerf's neural radiation field is composed of the hash voxel field in instance-NGP\cite{muller2022instant}, mapping the distance interval $T_i = [t_i,t_{i+1})$ on a ray $\bm{\mathrm{r}}(t) = \bm{\mathrm{o}} + t\bm{\mathrm{d}}$ into a set of color features $\bm{\mathrm{c}} \in [0,1]^3 $ and volume density $\sigma \in \mathbb{R}^+ $, where $\bm{\mathrm{o}}$ and $\bm{\mathrm{d}}$ are the origin and direction of the ray, respectively, $t$ is the distance from the origin along the ray direction. It can be formulated as:

\begin{equation}
	(\sigma_i,\bm{\mathrm{c}}_i) = \mathrm{MLP}_\theta(\gamma(\bm{\mathrm{r}}(T_i))) , \quad\forall T_i \in \bm{\mathrm{t}}
\end{equation}

Here, $\mathrm{MLP}_\theta$ is an shallow MLP with parameters $\theta$, $\bm{\mathrm{r}}(T_i)$ maps the sampling points in the conical frustum corresponding to $T_i$, $\gamma$ is the coding function, and $\bm{\mathrm{t}}$ is the set of all intervals on the ray that are included in the rendering.

\textbf{Volume rendering:} With these volume densities and colors, we can calculate the RGB values $\bm{\mathrm{C}}$ corresponding to the rays based on the volume rendering formula:

\begin{equation}
	\bm{\mathrm{C(r,t)}} = \sum_i w_i \bm{\mathrm{c}}_i
\end{equation}

\begin{equation}
	w_i=\left(1-e^{-\sigma_i\left(t_{i+1}-t_i\right)}\right) {E}_{i}
\end{equation}

\begin{equation}
	E_{i} = e^{-\sum_{j<i} \sigma_{j}\left(t_{j+1}-t_{j}\right)}
\end{equation}

Where $E_i$ is the accumulated transmittance along the ray from viewpoint to $t_i$, $\left(1-e^{-\sigma_i\left(t_{i+1}-t_i\right)}\right)$  is the opacity at $T_i$, By construction, the sum of the weights $w_i$ on a ray is always less than or equal to 1, and when the ray points to an opaque surface, the sum of the weights $w_i$ approaches 1 \cite{barron2022mip}. 
Usually, the scene depth calculation formula corresponding to a ray can be written as:
\begin{equation}
	z\bm{\mathrm{(r,t)}} = \sum_i w_i t_{mid}, \quad t_{mid}=\frac{(t_i+t_{i+1})}{2} 
	\label{eq:eq5}
\end{equation}   

Here, $z\bm{\mathrm{(r,t)}}$ is the rendering depth of ray $\bm{\mathrm{r}}$. As the sum of $w_i$ always approaches 1. We can also consider Eq (\ref{eq:eq5}) as calculating the expected value, and $w_i$ is the probability of the existence of an object surface at position $t_{mid}$.

\textbf{Midpoint depth:} 
Although most Nerf-based generative methods\cite{tosi2023nerf,truong2023sparf,yang2023unisim,yang2023emernerf} use the depth described in Eq (\ref{eq:eq5}), it is unreasonable in Zip-Nerf. As mentioned in the previous chapter, its definition is the expected distance between all surfaces on the ray. So, when there are multiple surfaces or incorrect surfaces on a ray (due to insufficient Nerf reconstruction), it is highly likely to generate errors.

In this paper, we use midpoint depth as the rendering depth output, which is defined as follows:
\begin{equation}
	z\bm{\mathrm{(r,t)}} = t_n, \quad   \sum_i^n w_i = 0.5
	\label{eq:eq61}
\end{equation} 

\begin{figure}[!t]
	\centering
	\includegraphics[width=2.4in]{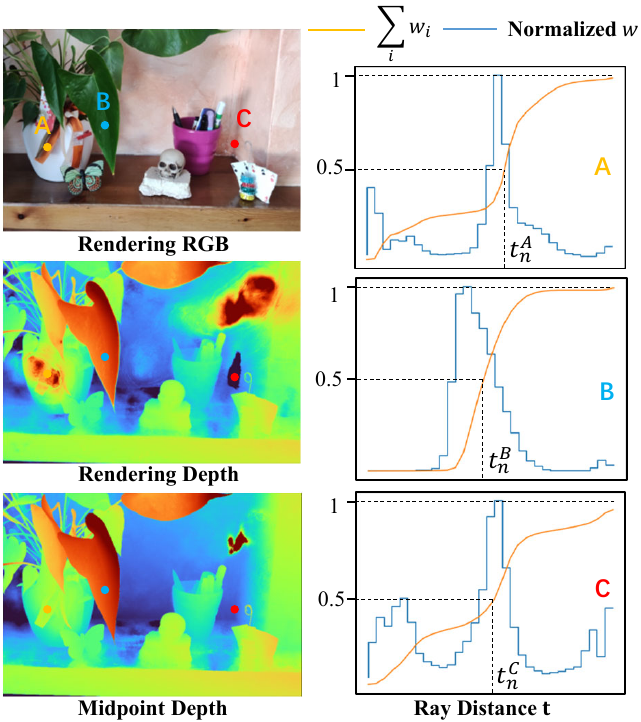}
	\caption{\textbf{Midpoint Depth vs. Rendering Depth.} Left: Rendered RGB image and two different depths. Right: The weights $w$ on three different rays: well-trained ray B, ray C with incorrect surfaces, and ray A with potential multiple surfaces.
		We found that rendering depth in A and C is clearly incorrect, as the weighted depth of the wrong surface interfered with the final result, especially when there were incorrect weights at the far end of the ray. 
		And the midpoint depth can reduce the interference of these erroneous surfaces.}
	\label{fig_1}
\end{figure}
\begin{figure*}[!t]
	\centering
	\includegraphics[width=6.6in]{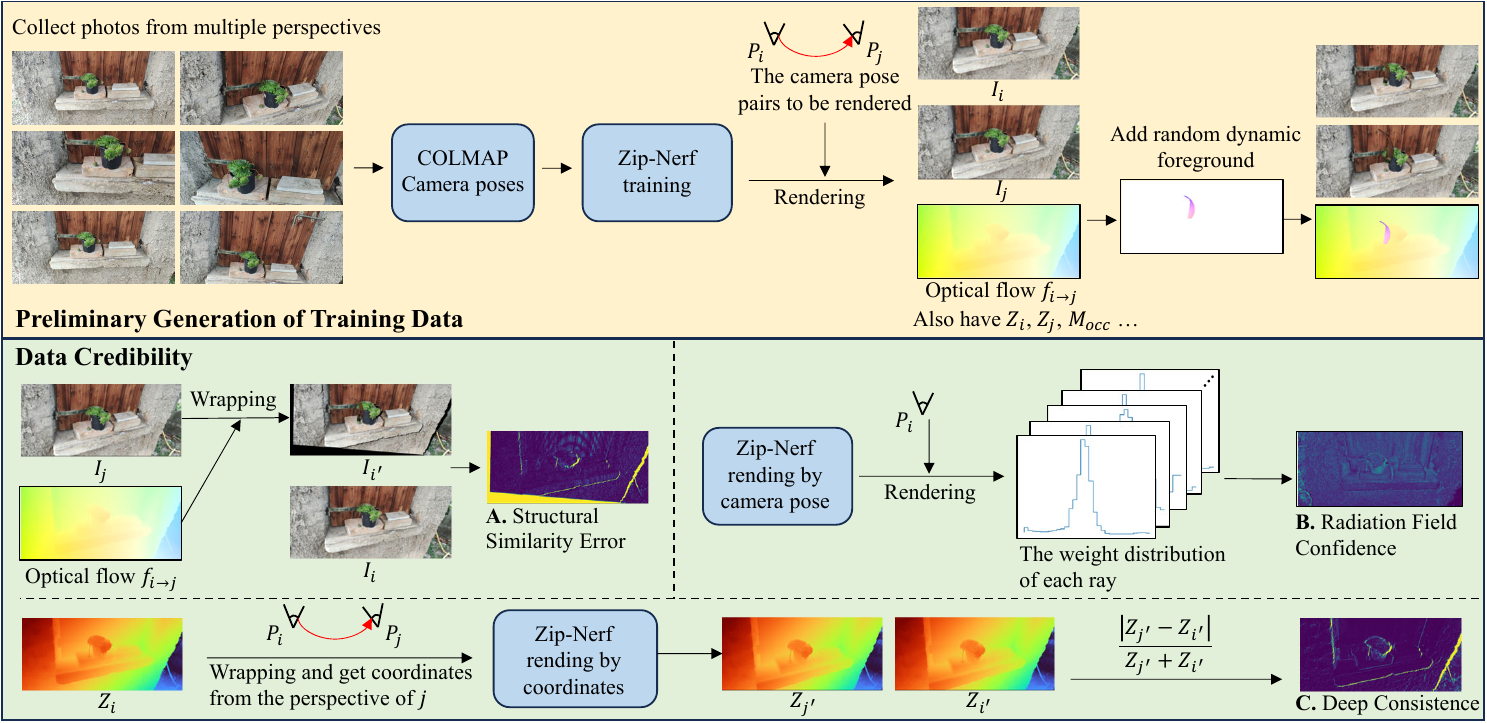}
	
	\caption{\textbf{Data Factory Workflow.} On the top: Training Nerf using a sequence of photos collected by a monocular camera and rendering preliminary optical flow labels. Below: Conduct confidence checks on the rendered data and labels, eliminating non-conforming parts. }
	\label{fig_2}
\end{figure*}
Due to the anti-aliasing characteristics of Zip-Nerf, the sum of weight $w$ on a well-trained ray should appear as a smoothed step function, as shown in Fig.\ref{fig_1} B. At this point, the midpoint distance is close to the peak of $w$ (Location with the highest potential surface probability). As shown in Fig.\ref{fig_1} A and C, when there are potential multiple surfaces or incorrect surfaces on the ray, the midpoint depth can also avoid small errors and approach the correct depth.

\section{Method}
Fig.\ref{fig_2} shows our data factory's workflow, which is divided into two stages. Firstly, we construct a neural radiation field based on static scene multi-view images users collected, and generate preliminary optical flow results. Then, we propose three metrics: structural similarity, radiation field confidence, and deep consistency to filter the generated data. The filtered optical flow results can be directly used for network training.
\subsection{Optical flow generation}
After giving the camera pose pair $P_i$ and $P_j$, we can obtain the RGB images $I_i, I_j$ and the corresponding depth $Z_i, Z_j$ based on Nerf rendering. Among them $I_i(u,v) = \bm{\mathrm{C(r,t)}}$, $Z_i(u,v) = z\bm{\mathrm{(r,t)}}$, $(u,v)$ is the pixel plane coordinate corresponding to ray $\bm{\mathrm{(r,t)}}$, $P$ is the $3 \times 4$ transformation matrix from the camera coordinate system to the world coordinate system.

Based on the $Z_i$ and the camera pose $P_i,P_j$, we can calculate the position of the pixel $p_i(u,v)=(u,v,1)$ in the first frame to corresponding $p_{i'}(u,v)=(u',v',1)$ in the second frame, as follows:

\begin{equation}
	p_{i'}(u,v) = \frac{Z_i(u,v) Kp_i(u,v) P_i {P_j}^{-1} K^{-1}}{Z_{i'}(u,v)}
	\label{eq:eq7}
\end{equation}   
\begin{equation}
	f_{i\rightarrow j} = p_{i'} - p_{i}
	\label{eq:eq8}
\end{equation}  

Where $ f_{i\rightarrow j}$ is the optical flow between frames $i$ and $j$, $K$ is the camera's internal reference matrix to convert pixel points into the camera coordinate system, $Z_{i'}(u,v)$ is the depth of the point in $I_i$ from the perspective of $I_j$.

\textbf{Dynamic foreground:} Because Zip-Nerf can only model static backgrounds. To make up for this weakness, we use Bezier curves to form randomly shaped slices and make perspective changes between two frames to enhance the foreground part of the generated data, as shown in Fig.\ref{fig_2}.

In addition to the optical flow results, based on $Z_{i}$ and $Z_{j}$, we can also calculate scene flow datasets. For more information on datasets and foreground masks, please refer to \textbf{supplementary materials}.
\subsection{Data Credibility}
During shooting, the environment may change and the shooting perspective may not be perfect, which can result in artifacts in the data generated by Nerf. To address this issue, we require a reliable filtering mechanism to screen out unqualified data. In previous studies\cite{truong2023sparf}, researchers have used ambient occlusion (AO) to measure the credibility of generated data. The AO can be calculated for a ray using the following formula:
\begin{equation}
	\mathrm{AO} = \sum_{i}^{n-1} w_i, \quad t_n = z\bm{\mathrm{(r,t)}}
	\label{eq:eq9}
\end{equation}   

The meaning of AO is the probability of the presence of surfaces from the rendering depth $z\bm{\mathrm{(r,t)}}$ to the viewpoint. However, $\mathrm{AO}$ is not applicable in anti-aliasing methods because $w$ has been smoothed, so even ordinary rays will have high $\mathrm{AO}$ values. In this article, we only use $\mathrm{AO}$ to calculate occlusion mask $M_{occ}$ between two frames. At this time, we use $Z_{i'}$ in Eq.\ref{eq:eq7} to replace the rendering depth. The specific calculation process is as follows:

\begin{equation}
	\mathrm{AO_{occ}} = \sum_{j}^{n-1} w_j, \quad t_n = Z_{i'}(u,v)
	\label{eq:eqocc}
\end{equation}  
\begin{equation}
	M_{occ}(u,v)= \begin{cases}0 & \text { if } \mathrm{AO_{occ}}<t h \\ 1 & \text { otherwise }\end{cases}
\end{equation}

Where $\mathrm{AO_{occ}}$ is the probability of the presence of a surface between the viewpoint and the depth $Z_{i'}(u,v)$, $M_{occ}$ is the occlusion mask between two frames, and in experiments, $th$ is generally set to 0.3.

In the following section, we propose different credibility masks to filter the generated data of Nerf from three perspectives: optical flow matching accuracy, Nerf rendering stability, and depth consistency. Combining these indicators can find generated data labels that conform to geometric relationships.
\begin{figure}[!t]
	\centering
	\includegraphics[width=2.8in]{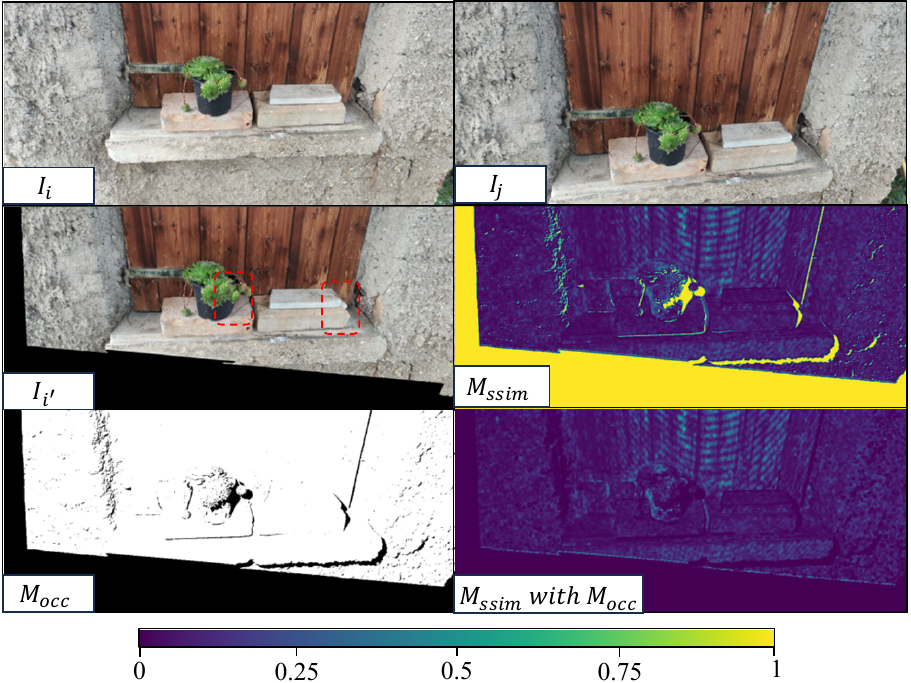}
	\caption{\textbf{SSIM Inspection of Optical Flow Labels.} We got $I_{i'}$ by warping $I_j$ to $I_i$ based on optical flow, and calculated the $M_{ssim}$ between $I_{i'}$ and $I_{i}$, where the darker the color, the smaller the error. Notice that the part in the red box has artifacts due to occlusion. When using $M_{ssim}$ as the evaluation standard, we filtered out these artifacts using an occlusion mask $M_{occ}$ to reflect the matching quality of the optical flow truly.}
	\label{fig_3}
\end{figure}

\textbf{Optical flow matching accuracy:}
As shown in Fig.\ref{fig_3}, to evaluate the accuracy of generating optical flow $ f_{i\rightarrow j}$, we used the optical flow results to warp the second frame image $I_j$ and depth $Z_j$ to the first frame to obtain $I_{i'}$. Then we calculated the corresponding structural similarity masks:
\begin{equation}
	M_{ssim} =1- \mathrm{SSIM}(I_{i},I_{i'})
	\label{eq:eq10}
\end{equation}

where $ \mathrm{SSIM}$\cite{
	wang2004image} is the Structural Similarity Index Measure. 
$M_{ssim}$ can also be used as an evaluation indicator of optical flow. In this case, it needs to be used in conjunction with the $M_{occ}$ mask to eliminate artifacts caused by occlusion, as shown in Fig.\ref{fig_3}.

\begin{figure}[!t]
	\centering
	\includegraphics[width=2.8in]{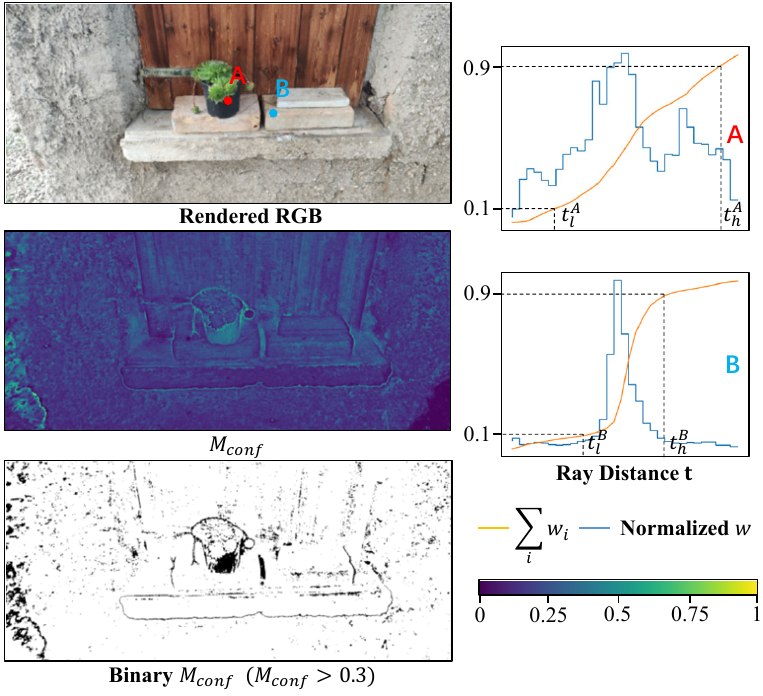}
	\caption{\textbf{Visualization of Radiation Field Confidence.} Left: RGB image rendered by neural field, $M_{conf}$ confidence mask, and binarized $M_{conf}$. Right: A represents the weight distribution of rays in dark areas without texture, and B represents the weight distribution of rays in rich texture areas. The $M_{conf}$ we proposed easily segmented those areas where the radiation field is difficult to reconstruct based on the ray weight distribution.}
	\label{fig_4}
\end{figure}

\textbf{Radiation field confidence:}
In Nerf generative application scenarios, the quality of neural field reconstruction directly determines the method's overall performance. However, the only available AO indicators are not widely applicable for various methods and scenarios. In order to overcome the above difficulties, we propose a new metric in this section to measure the quality of neural field reconstruction, called radiation field confidence (RFC).

As shown in Fig.\ref{fig_4}B, when the neural field is well trained, the weight distribution of rays should be clustered at a central point, and the weight integration should appear as a step function, indicating that the radiation field has an exact surface on that ray. Moreover, we observed that such rays often correspond to areas with rich textures and appropriate brightness. On the contrary, it is difficult for the radiation field to learn the correct surface position for areas that are too dark or textureless.

Based on the above observations, we proposed $M_{conf}$ (RFC) to describe the reconstruction quality of the neural field, as follows:

\begin{equation}
	M_{conf}(u,v) = \frac{t_h - t_l}{t_h + t_l}
	\label{eq:eq12}
\end{equation}   
\begin{equation}
	\sum_i^l w_i = th_{low}, \quad \sum_i^h w_i = th_{high}
	\label{eq:eq13}
\end{equation}  

Where $th_{low}$ and $th_{high}$ are the boundary thresholds, which are set to 0.1 and 0.9 in the experiments of this paper, $t_l$ and $t_h$ are the depths in the ray direction when the boundary threshold is reached.

 $M_{conf}$ revealed the dispersion of weight distribution on a ray, with a range of $(0,1)$. The larger the value of $M_{conf}$, the lower the reconstruction quality of the radiation field. As shown in Fig.\ref{fig_4}, $M_{conf}$ can effectively identify the parts of where the radiation field is challenging to reconstruct.

\textbf{Geometry consistency:}
Inspired by Bian\cite{bian2021unsupervised}, we calculated the geometric consistency mask of rendering depth under multiple perspectives. This is a core component for scene flow and depth estimation tasks. Although this paper focuses on optical flow applications, we hope the data factory proposed can be further applied to a broader range of downstream tasks.

Specifically, as shown in Fig.\ref{fig_2}, given the rendering depth $Z_i$ of the first frame and the pose $P_i,P_j$ between the two perspectives. 
We can calculate the coordinates $p_{i'}$ and depth map $Z_{i'}$  in $P_j$ corresponding to $P_i$ based on Eq.\ref{eq:eq7}, and then use Nerf rendering to obtain the depth value $Z_{j'}$ of the corresponding coordinates  $p_{i'}$. The final geometric consistency is as follows:

\begin{equation}
	M_{dc} =\frac{\left| Z_{j'} - Z_{i'}\right|}{ Z_{j'} + Z_{i'}}
	\label{eq:eq16}
\end{equation}

 Where the value range of $M_{dc}$ is (0,1). The larger the value of $M_{dc}$, the lower the depth consistency.

 \subsection{NeRF-Supervised Training }
 After calculating the optical flow value $f_{i\rightarrow j}$ between two frames, we filter the optical flow results based on the mask $M_{conf}$, $M_{ssim}$, and $M_{dc}$. In this paper, we simply set a threshold to filter out those unqualified points, and the final optical flow label $f_{i\rightarrow j}^{gt}$ used for training is as follows:
 
 \begin{equation}
 	f_{i\rightarrow j}^{gt} = M_{conf}^{th1}M_{ssim}^{th2}M_{dc}^{th3}f_{i\rightarrow j}
 	\label{eq:eq166}
 \end{equation}   
\begin{equation}
	M^{th}= \begin{cases}1 & \text { if } M <t h \\ 0 & \text { otherwise }\end{cases}
\end{equation}

Points with $f_{i\rightarrow j}^{gt}$ equal to zero do not participate in the loss calculation.

\begin{table*}[htbp]
	\centering
	\caption{\textbf{Ablation Study.} The best results in ablation studies are bolded. $Fl_{epe}$ is the average end-to-end optical flow error, $Fl_{all}$ is the optical flow outlier rate (errors greater than 3 pixels or greater than 5\% are considered outliers) }
	\setlength{\tabcolsep}{4.8pt}
	\begin{tabular}{llccccccccccc}
		\toprule
		&       &       &       &       &       &       & \multicolumn{2}{c}{K15} & \multicolumn{2}{c}{Midd-A} & \multicolumn{2}{c}{K12} \\
		\midrule
		Method & \multicolumn{1}{l}{Training Data} & $M_{conf}$ & $M_{ssim}$ & $M_{dc}$   & $M_{occ}$  & $M_f$ & $Fl_{epe}$  & $F_{all}$ & $Fl_{epe}$   & $F_{all}$ & $Fl_{epe}$   & $F_{all}$ \\
		\midrule
		\multicolumn{1}{l|}{\multirow{10}[4]{*}{RAFT\cite{teed2020raft}}} &       &       &       &       &       &       & 4.9   & 17.34 & 0.313 & 0.222 & 2.04  & 9.18 \\
		\multicolumn{1}{c|}{} & \multicolumn{1}{l}{\multirow{6}[1]{*}{ADF15 (ours)}} &       &       &       &       & \checkmark & 4.62  & 15.71 & 0.368 & 0.184 & 1.81  & 7.45 \\
		\multicolumn{1}{c|}{} &       &       &       &       & \checkmark & \checkmark & 4.56  & 19.2  & 0.281 & 0.169 & 1.99  & 11.55 \\
		\multicolumn{1}{c|}{} &       &       &       & \checkmark &       & \checkmark & 4.42  & 18.28 & \textbf{0.201} & 0.186 & 1.77  & 9.59 \\
		\multicolumn{1}{c|}{} &       &       & \checkmark &       &       & \checkmark & 4.4   & 15.7  & 0.257 & \textbf{0.129} & 1.72  & 7.76 \\
		\multicolumn{1}{c|}{} &       & \checkmark &       &       &       & \checkmark & 4.19  & 14.83 & 0.3   & 0.231 & 1.73  & 8.12 \\
		\multicolumn{1}{c|}{} &       & \checkmark & \checkmark & \checkmark & \checkmark & \checkmark & \textbf{4.19} & \textbf{14.15} & 0.28  & 0.175 & \textbf{1.62} & \textbf{6.8} \\
		\cmidrule{2-13}    \multicolumn{1}{c|}{} & \multicolumn{1}{l}{Sintel} &       &       &       &       &       & 8.5   & 18.31 & 0.23 & 0.22  & 2.38  & 8.01 \\
		\multicolumn{1}{c|}{} & \multicolumn{1}{l}{Flyingthings3D} &       &       &       &       &       & 8.2   & 23.42 & 0.24  & 0.17  & 3.14  & 14.16 \\
		\multicolumn{1}{c|}{} & \multicolumn{1}{l}{FlyingChairs2} &       &       &       &       &       & 10.87 & 36.86 & 0.46  & 0.79  & 4.2   & 26 \\
		\midrule
		Scale-flow\cite{ling2022scale} & ADF58 (ours) & \checkmark & \checkmark & \checkmark & \checkmark & \checkmark & 3.88  & 13.36 & 0.218 & 0.122 & 1.59  & 6.97 \\
		RAFT & ADF58 (ours) & \checkmark & \checkmark & \checkmark & \checkmark & \checkmark & 4.17  & 13.9 & 0.223 & 0.138 & 1.59  & 6.43 \\
		\bottomrule
	\end{tabular}%
	\label{tab:tab1}%
\end{table*}%

\section{Experimental}
In this section, we first introduce the implementation details of the dataset and experiment and then discuss the experimental results.

\textbf{Generate dataset}
We have gathered and created a series of 300 scenes, featuring both indoor still life and outdoor courtyard scenes and ensured complete stillness. For each scenario, we train for 30,000 steps with a batch size of 10240 based on Zip-Nerf. Afterwards, we generate approximately 200 camera pose pairs per scene (maintaining orientation towards the center of the scene) to produce an optical flow dataset and corresponding masks. In the end, the training set consists of a total of 58800 image pairs (ADF58).


\textbf{Deep optical flow training:}
We use RAFT\cite{teed2020raft} as the main architecture for ablation experiments and optical flow experiments evaluation, as it is the backbone network of most advanced methods. Hence, the evaluation on it has better representativeness. Of course, we also trained a stronger normalized scene flow baseline Scale-flow\cite{ling2022scale} to evaluate the effectiveness of ADF in more advanced methods.
All methods are trained with a batch size of 6 and a crop size of 384×768, where we trained 200k iterations for RAFT and 300k iterations for Scale-flow. All training is done from scratch without using any synthesized dataset for pre-training.

\textbf{Datasets:} In the evaluation, we used 194 images from KITTI 2012 training set\cite{geiger2012we} (K12), 3 static scenes from Middlebury\cite{baker2011database} (Midd-A), and 200 images from KITTI 2015 training set\cite{Menze2015ISA} (K15).
Around 30k KITTI raw data (Kr) and 4k KITTI multi-frame (Km) images were used for training in other self-supervised methods\cite{kong2022mdflow,luo2021upflow}.
In addition, commonly used optical flow pre-trained synthetic datasets Sintel\cite{butler2012naturalistic} (S), Flyingthings3D\cite{mayer2016large} (T), and Flyingchair2\cite{mayer2016large} (C2) were also included in the comparison. To facilitate the ablation experiment, we separated a subset of 15600 images (ADF15) from the complete ADF58.

\subsection{Ablation Study}

In this section, we first studied the effectiveness of various masks proposed in this article. Then, we compared the datasets ADF58 and ADF15 produced by our data factory with the commonly used synthetic dataset. Except for ADF58, all datasets only train 100K iterations.

\textbf{Use of masks:} We simply set a threshold to filter out unqualified points.
Specifically, points with $M_{conf}$ greater than 0.3 will be excluded, those with  $M_{ssim}$ greater than 0.1 will be excluded, those with  $M_{occ}$ greater than 0.3 will be regarded as occlusion, and those with $M_{dc}$ greater than 0.01 will be excluded.
In our experiments, the number of added motion foreground $M_f$ is 2.

\textbf{Mask analysis:}
Tab.\ref{tab:tab1} shows the experimental results of training RAFT using different masks. Firstly, we found that the training effect after using masks always improves in most cases. Specifically, after adding $M_f$ as a sports foreground, the performance in the KITTI dataset was significantly improved, but there was a decrease in Midd-A. This is easy to understand because $M_f$ enhances the network's perception of motion prospects, while Midd-A is all static scenes. Secondly, we found that $M_{conf}$ had the most significant performance improvement among all masks, which proves that our motivation to evaluate dataset quality from the perspective of reconstruction stability is effective.
Finally, RAFT achieved optimal results after using all masks, which proves that the various masks we proposed are compatible. Although this article only proves its effectiveness by applying masks through simple thresholding, it is necessary to use masks more finely in subsequent work, such as calculating the credibility of points by combining masks in different tasks.


\textbf{Comparison with synthetic datasets:}
For fairness, we compared ADF15 with the commonly used synthetic datasets Sintel, Flyingthings3D and Flyingchairs2. All of them were trained using the same training parameters based on RAFT. Tab.\ref{tab:tab1} shows the comparison results, and ADF15 is far ahead in almost all indicators, thanks to its data domain being closer to the KITTI scenario. It has been proven that our Nerf data factory scheme can serve as a substitute and supplement to existing synthetic datasets for pre-training matching tasks.

\textbf{Number of scenes and stronger baseline:} 
We attempted to use more scenarios for training, and as expected, ADF58 improved all metrics compared to ADF15, representing a key advantage of our work: we can quickly expand the photos taken casually into the dataset and enhance the method's performance.
In addition, we also tested a stronger normalized scene flow method, Scale-flow, and achieved better performance than RAFT, which proves that our method is also effective on a stronger baseline.
\begin{table}[htbp]
	\centering
	\caption{\textbf{Evaluation of Normalized Scene Flow.}  Above is the depth method that only uses monocular images, and below is the traditional method that uses stereo images.}
	\setlength{\tabcolsep}{3.6pt}
	\begin{tabular}{lcccc}
		\toprule
		Method & \multicolumn{1}{l}{Training data} & $Fl_{epe}$   & $Mid_{error}$   & \multicolumn{1}{l}{Time/s} \\
		\midrule
		Scale-flow & ADF58(ours) & \textbf{3.46} & \textbf{149.83} & 0.2 \\
		Expansion\cite{yang2020upgrading} & Kr & 6.56  & 348   & 0.2 \\
		SMMSF\cite{hur2021self} & Km+Kr & 7.92  & 288.99 & \textbf{0.063} \\
		\midrule
		OSF\cite{menze2018object}   & K15 &   -   & \textbf{115} & \textbf{300} \\
		PRSM\cite{vogel20153d}  & K15 &   -   & 124   & 3000 \\
		\bottomrule
	\end{tabular}%
	\label{tab:tab2}%
\end{table}%

\subsection{ Normalized Scene Flow }

This section compares the Scale-flow trained on ADF58 with other monocular self-supervised scene flow methods. For fairness, we use the same evaluation criteria as Yang\cite{yang2020upgrading}. Specifically, using 40 images in K15 for evaluation, the metrics include depth change rate $\tau$ and end-to-end optical flow error $Fl_{epe}$. The calculation of depth change rate loss $Mid_{error}$ is as follows:
\begin{equation}
	Mid_{error} = ||log(\tau)-log(\tau_{GT})||_1\cdot10^4
	\label{eq:eq15}
\end{equation}

Where $\tau = {z_2}/{z_1}$, $z_1$ is the depth of point in the first frame, and $z_2$ is the depth of the corresponding point  in the second frame.

We first compared our method with the state-of-the-art monocular self-supervised scene flow method SMMSF\cite{hur2021self}. Our Scale-flow with ADF58 outperformed them by a large margin(149.83 vs. 288.99). Because too complex models often have greater training difficulty, the architecture of general unsupervised methods tends to favour models with small parameter numbers, which is also why SMMSF is smaller and faster. Unlike them, our data factory solution can fully train more complex and more extensive models.

We also compare with traditional methods\cite{menze2018object,vogel20153d} that use stereo images as input. They decompose the image into rigid blocks, and iteratively optimize the 3D motion of the rigid blocks based on the rigid assumption and regularization terms. As shown in Tab.\ref{tab:tab2}, we have achieved accuracy close to theirs, and the speed is much faster (0.2s vs. 300s).

\subsection{Self-supervised Optical Flow}
In this section, we compare with several state-of-the-art self-supervised optical flow methods: MDFlow\cite{kong2022mdflow}, UPFlow\cite{luo2021upflow}, and UFlow\cite{jonschkowski2020matters}. Because MDFlow only provides fast mode weights, the MDFlow in our paper defaults to MDFlow-fast.
\begin{table}[htbp]
	\centering
	\caption{\textbf{Evaluation of Self-supervised Optical Flow.} Above is the zero-shot generalization result, and below is the result of fine-tuning using target domain data.}
	\setlength{\tabcolsep}{3.2pt}
	\begin{tabular}{lccccc}
		\toprule
		&       & \multicolumn{2}{c}{K15} & \multicolumn{2}{c}{K12} \\
		\midrule
		Method & Training data & $Fl_{epe}$   & $F_{all}$  & $Fl_{epe}$   & $F_{all}$ \\
		\midrule
		MDFlow\cite{kong2022mdflow} & Sintel & 10.05 & 23.12 & 3.49  & 12.17 \\
		MDFlow\cite{kong2022mdflow} & GTA5  & 9.13  & 25.01 & 3.85  & 14.33 \\
		Scale-flow & ADF58(ours) & \textbf{3.89} & \textbf{13.57} & \textbf{1.59} & 6.97 \\
		RAFT  & ADF58(ours) & 4.19  & 14.15 & 1.62  & \textbf{6.8} \\
		\midrule
		UFlow\cite{jonschkowski2020matters} & Km+Kr & 2.71  &   -  & 1.68  & -\\
		MDFlow\cite{kong2022mdflow} & Km  & 4.44  & \textbf{12.3}  & 1.83  & \textbf{6.8} \\
		UPFlow\cite{luo2021upflow} & Km+Kr & \textbf{2.45} &   -   & \textbf{1.27} &  \\
		SMMSF\cite{hur2021self} & Km+Kr & 6.04  & 18.81 &    -  &  -\\
		\bottomrule
	\end{tabular}%
	\label{tab:tab3}%
\end{table}%

As shown in the upper part of Tab.\ref{tab:tab3}, our method has achieved absolute advantages over the self-supervised algorithm in the case of zero-shot generalization. However, after fine-tuning the self-supervised method on the target domain data, the difference between self-supervised and our zero-shot generalization results quickly decreased or even exceeded, as shown in the lower part of  Tab.\ref{tab:tab3}.

This experiment exposes a potential weakness of the data factory solution. In practical use, it requires manual capture of static scenes in the target data domain, so it is hard to build fine-tuning datasets on KITTI directly.

\subsection{Zero-shot Generalization in Real Word}
Finally, we test our algorithm in real-world daily scenarios, which is also the original intention of this work: to enable the massive number of excellent optical flow algorithms and scene flow algorithms to be truly applied to our daily production and life.

\textbf{Evaluation method:} As there is currently no large-scale real-world optical flow dataset, we have adopted an indirect evaluation method to evaluate optical flow accuracy based on SSIM and photometric loss.

Assuming $f$ is the optical flow between frames $I_i$ and $I_j$, in order to evaluate its accuracy, we project $I_j$ frames onto $I_i$ frames based on $f$ to obtain ${I_i}'$.The specific calculation method is as follows:

\begin{equation}
	S_{loss}= M_{occ} \cdot M_{ssim}
	\label{eq:eq16}
\end{equation}   
\begin{equation}
	P_{loss}= M_{occ} \cdot \left|I_{i}-I_{i'}\right|
	\label{eq:eq17}
\end{equation} 

Where $S_{loss}$ and $P_{loss}$ represent SSIM loss and photometric loss, respectively, while $M_{occ}$ represents the occlusion mask between two frames, used to eliminate artifacts caused by occlusion.

\textbf{Evaluation dataset:} We chose DAVIS\cite{pont20172017} as the evaluation dataset because it contains many dynamic prospects and a wide distribution of data domains. Moreover, it has a foreground mask, which facilitates the calculation of occluded parts. Finally, approximately 3,400 image pairs were included in the evaluation.
\begin{table}[htbp]
	\centering
	\caption{Evaluation of Zero-shot Generalization in Real-world. ${fg}$ and ${bg}$ represent the foreground and background.}
	\setlength{\tabcolsep}{4.0pt}
	\begin{tabular}{lccccc}
		\toprule
		Method & Training data & $S_{loss}^{fg} $ & $S_{loss}^{bg} $ & $P_{loss}^{fg}$ & $P_{loss}^{bg}$ \\
		\midrule
		Scale-flow & ADF58(ours) & \textbf{0.21} & \textbf{0.09} & \textbf{11.56} & \textbf{4.13} \\
		RAFT  & ADF58(ours) & 0.22 & 0.09 & 11.69 & 4.19 \\
		GMFlow\cite{xu2022gmflow} & KITTI-ft & 0.31 & 0.15 & 15.81 & 5.97 \\
		Scale-flow & KITTI-ft & 0.26 & 0.10 & 13.62 & 4.55 \\
		RAFT  & KITTI-ft & 0.29 & 0.11 & 15.15& 4.82 \\
		\bottomrule
	\end{tabular}%
	\label{tab:tab6}%
\end{table}%

We mainly compare the performance differences under different supervision methods. As shown in Tab.\ref{tab:tab6}, KITTI-ft represents the commonly used fine-tuning training process, which involves pre-training on extensive synthetic data such as Sintel and Flyingthings3D, then fine-tuning on the real-world dataset KITTI. ADF58 represents our self-supervised data factory solution. RAFT and Scale-flow exhibits a significant difference in generalization ability under different training methods (0.21 vs. 0.26; 0.22 vs. 0.29). This proves that our data factory scheme has better generalization performance in real-world scenarios compared to the previous supervised training mode.

\begin{figure}[!t]
	\centering
	\includegraphics[width=3.3in]{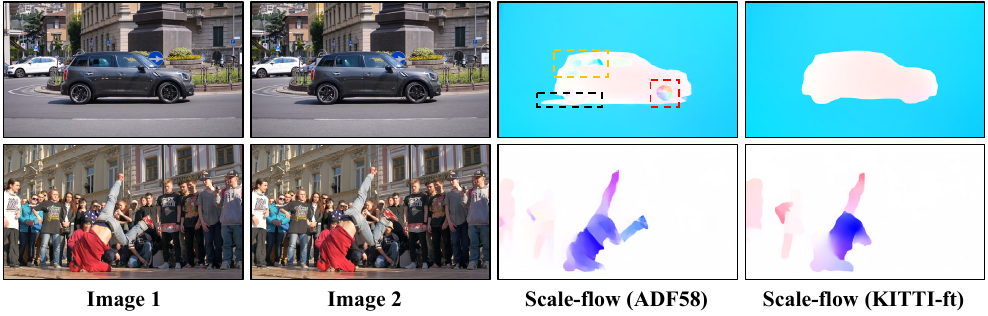}
	\caption{\textbf{Optical Flow and Object Flow:}. Observing the parts in coloured boxes, the methods trained by ADF are more faithful to the flow of light, while existing supervised methods (KITTI-ft) often tend to learn object flows with semantics. In addition, due to the lack of training samples, supervised methods often have poorer generalization in the real world.}
	\label{fig_6}
\end{figure}

\section{Limitation and Improvement}
In this section, we discuss the limitations and potential improvement methods of our method.

\textbf{Optical flow and object flow:}
Observing the car in Fig.\ref{fig_6}, we found that our method has some more details compared to the supervised learning method. For example, the rotating wheels in the red box, neglected transparent glass in the orange box, and shadow in the black box. This is because our method is based on Nerf, so the results are more faithful to the light flow. In existing supervised learning\cite{Menze2015ISA}, optical flow is defined as object flow, which tracks the surface of objects in space and introduces semantic information to erase some details, such as rotating car wheels and transparent windows. In future work, we believe that we can try to bring our data factory closer to supervised learning by introducing a big semantic model\cite{li2023semantic}.

\textbf{Nerf:}
The Zip-Nerf used in this paper, cannot reconstruct reflective surfaces such as glass and water surfaces, as well as limiting to static scenes. These limitations directly prevent us from creating fine-tuning scenarios based on KITTI raw data.
Fortunately, thanks to the rapid development of Nerf technology\cite{verbin2022ref,attal2021torf,luiten2023dynamic,yang2023emernerf,wu2022d}, the problem has been partially resolved.

\section{Conclusion}
We propose a groundbreaking optical flow training method: Automated Data Factory (ADF), which utilizes scenes generated by Nerf to train deep optical flow networks without requiring manual annotation or expensive equipment. ADF can generate almost infinite high-quality optical flow datasets from ordinary monocular cameras,  and directly use them for network training. This approach results in excellent real-world zero generalization performance, surpassing most self-supervised and supervised methods. Our work can help many excellent optical flow algorithms perform better in the real world, making them applicable to production and life!

{
    \small
    \bibliographystyle{ieeenat_fullname}
    \bibliography{main}
}


\end{document}